\pdfoutput=1

\documentclass[11pt]{article}

\usepackage[preprint]{acl}

\usepackage{times}
\usepackage{latexsym}

\usepackage[T1]{fontenc}

\usepackage[utf8]{inputenc}

\usepackage{microtype}

\usepackage{inconsolata}

\usepackage{graphicx}
\usepackage{booktabs}
\usepackage{xcolor}
\usepackage[normalem]{ulem}
\useunder{\uline}{\ul}{}
\usepackage{tabularx}
\usepackage{amsmath}
\newcolumntype{Y}{>{\centering\arraybackslash}X}

\title{What’s Taboo for You?\\ An Empirical Evaluation of LLMs Behavior Toward Sensitive Content}

\author{
\textbf{Alfio Ferrara\textsuperscript{1}},
\textbf{Sergio Picascia\textsuperscript{1}},
\textbf{Laura Pinnavaia\textsuperscript{2}},\\
\textbf{Vojimir Ranitovic\textsuperscript{3}},
\textbf{Elisabetta Rocchetti\textsuperscript{1}},
\textbf{Alice Tuveri\textsuperscript{2}},
\\
\textsuperscript{1}Università degli Studi di Milano, Department of Computer Science, Via Celoria, 18 - 20133 Milan, Italy\\
\textsuperscript{2}Università degli Studi di Milano, Department of Languages, Literatures, Cultures and Mediations,\\ Piazza S. Alessandro, 1 - 20123 Milan, Italy\\
\textsuperscript{3}Università degli Studi di Milano, Department of Historical Studies, Via Festa del Perdono, 7 - 20126 Milan, Italy\\
\small{
\textbf{Correspondence:} name.surname@unimi.it
}
}

\begin{document}
\maketitle

\begin{abstract}
Proprietary Large Language Models (LLMs) have shown tendencies toward politeness, formality, and implicit content moderation. While previous research has primarily focused on explicitly training models to moderate and detoxify sensitive content, there has been limited exploration of whether LLMs implicitly sanitize language without explicit instructions. This study empirically analyzes the implicit moderation behavior of GPT-4o-mini when paraphrasing sensitive content and evaluates the extent of sensitivity shifts. Our experiments indicate that GPT-4o-mini systematically moderates content toward less sensitive classes, with substantial reductions in derogatory and taboo language. Also, we evaluate the zero-shot capabilities of LLMs in classifying sentence sensitivity, comparing their performances against traditional methods.
\end{abstract}

\color{red}
\textbf{Disclaimer}: This paper includes examples of sensitive and very offensive language solely to illustrate the behavior of LLMs navigating sensitive language.

\color{black}

\section{Introduction}
\label{sec:intro}
Recent progress in the development of large language models (LLMs) has substantially transformed the way humans and machines communicate. These systems understand context and even perceive emotional cues, making conversations with machines feel much more natural and human~\citep{human_gpt_communication}.
One evident characteristic of these models is their tendency toward politeness, formality, and content moderation (unless prompted otherwise). Users interacting with ChatGPT often notice responses characterized by respectfulness and refinement, regardless of the tone used in the user's prompt. Besides outright refusals, usually triggered by inputs that violate OpenAI’s community guidelines and policies\footnote{\url{https://openai.com/policies/usage-policies}}~\citep{openai2024gpt4technicalreport}, there is an indication that proprietary GPT-based models subtly sanitize or moderate content even when no explicit violation occurs. 

State-of-the-art LLMs are trained using alignment techniques (e.g., Reinforcement Learning from Human Feedback, RLHF) to follow ethical guidelines and avoid harmful language~\citep{NEURIPS2022_b1efde53}. With that in mind, if such a model is asked to paraphrase text containing sensitive content, it might instinctively sanitize the output by, for example, removing slurs, aggressive tones, or profanity. While this kind of transformation could be highly beneficial from a content moderation perspective, limited research has quantified this implicit sanitization effect during paraphrasing. 

We empirically investigate and quantify content moderation within the context of paraphrasing. Specifically, we examine whether LLMs autonomously sanitize content during text generation and assess the extent to which this occurs through expert human judgment. Additionally, we explore how well LLMs can replicate human judgment in identifying problematic content. To this end, we set up an annotation task that classifies content into four sensitivity categories, listed here from least to most sensitive: \textit{Formal/Polite}, \textit{Informal}, \textit{Derogatory}, and \textit{Taboo}. We compare human expert annotations with the zero-shot classification output from LLMs to evaluate how closely LLMs’ perception of sensitivity aligns with that of humans. Furthermore, we trained traditional text classifiers on the expert annotations and compared their performance to that of the LLMs, providing an extra baseline for evaluating LLMs performance. This evaluation also offers the added benefit of determining whether a lightweight local model or traditional methods can reliably predict the same categories and detect sensitivity shifts as effectively as human annotators, potentially serving as a cost-effective alternative for monitoring LLM behavior at scale~\citep{Kumar2024-ad, multi-text-detox}.

In summary, our study is driven by the following research questions (RQ):
\begin{enumerate}
    \item[(RQ1)] Do proprietary LLMs perform implicit content moderation during paraphrasing, and if so, how significant are the changes in sensitivity compared to the original content?
    \item[(RQ2)] How closely do LLMs’ sensitivity perceptions align with humans, and how effectively can automated methods replicate human annotations and detect changes in sensitivity during paraphrasing?
\end{enumerate}

The paper is structured as follows: Section~\ref{sec:rel_work} reviews research on LLM alignment, moderation, and their use in content moderation and text detoxification; Section~\ref{sec:methodology} outlines our experimental design, including dataset construction, human annotations and automated classification, and evaluation methods; Section~\ref{sec:results} presents our findings; Section~\ref{sec:limitation} discusses the limitations of this work; Section~\ref{sec:conclusion} offers final thoughts.

\section{Related Work} 
\label{sec:rel_work}
Our work sits at the intersection of alignment, content moderation, and text refinement in LLMs, examining how an aligned paraphrasing model navigates sensitive language.

\paragraph{Alignment and moderation in LLMs.} As LLMs gain popularity, researchers have focused on techniques to reduce harmful outputs. OpenAI uses fine-tuning and reinforcement learning from human feedback (RLHF,~\citet{NEURIPS2022_b1efde53}) to mitigate toxicity and bias, with GPT-4 showing an 82\% reduction in disallowed content compared to GPT-3.5 and a significant drop in toxic output from 6.48\% to 0.73\% on the RealToxicityPrompts dataset~\citep{openai2024gpt4technicalreport, gehman2020realtoxicityprompts}. In contrast, Anthropic’s ``Constitutional AI'' method employs NLP rules for self-critiquing and revising responses, improving harmlessness without heavy human annotations~\citep{bai2022constitutionalaiharmlessnessai}. These approaches enable models to act as their own content moderators, either rejecting or modifying harmful outputs.

\paragraph{LLMs as content moderators.} Recent research has also evaluated LLMs as content moderators on user-generated text~\citep{gilardi2023,Kumar_AbuHashem_Durumeric_2024, Vargas_Penagos2024-ac}, leveraging their understanding of context. In particular,~\citet{Vargas_Penagos2024-ac} explores whether LLMs can automate decisions about what online content should be removed or allowed. It emphasizes that content moderation is not solely about removing illegal posts, but also about handling ``lawful but awful'' content, material that may be offensive yet legally protected. This poses unique challenges in balancing users’ rights with the need to maintain a safe public discourse. 

\paragraph{LLMs for text detoxification and politeness transfer.} Style transfer for polite rephrasing (also known as text detoxification) focuses on transforming text to remove or reduce offensive content while preserving meaning. One approach combines paraphrasing with style control to eliminate toxicity, such as using T5~\citep{t5} for paraphrasing and GPT-2~\citep{radford2019language} to replace toxic words~\citep{dale-etal-2021-text}. They also explore methods where BERT~\citep{devlin-etal-2019-bert} is used to substitute offensive words with neutral synonyms. ~\citet{som-etal-2024-demonstrations} proposes an in-context learning approach to offensive content paraphrasing, noting that paraphrasing can be a preferable alternative to the outright removal of toxic posts. ~\citet{tacl_a_00027} demonstrates that a model can generate responses using a politeness classifier in a feedback loop during generation. Tag-and-generate approaches have been proposed to transform rude sentences into polite ones~\citep{madaan-etal-2020-politeness}. These studies typically use parallel data or carefully curated datasets to ensure that the content remains the same while the style changes.

Our work empirically examines text sanitization and shifts in sensitivity without relying on explicit detoxification instructions, as seen in prior research. This allows us to observe potential sensitivity changes driven by the implicit style transfer of LLMs. Additionally, our approach is unique in that the parallel data is generated implicitly by the LLM itself, which paraphrases a selection of sentences.

\section{Experimental Design}
\label{sec:methodology}

In this section, we present the process designed for studying the behavior of LLMs toward sensitive language, exploiting their paraphrase and zero-shot classification capabilities. Firstly, Section~\ref{sec:data} discusses the collection of the original sentences and their paraphrases generated by an LLM; then, Section~\ref{sec:annotation} examines how sentences have been annotated, according to pre-defined sensitivity classes, by both human experts and automated approaches; finally, Section~\ref{sec:eval}, we explain the metrics and tools employed for interpreting the outcomes.


\subsection{Dataset Creation}
\label{sec:data}

\begin{figure*}[ht!]
  \centering
  \includegraphics[width=0.9\linewidth]{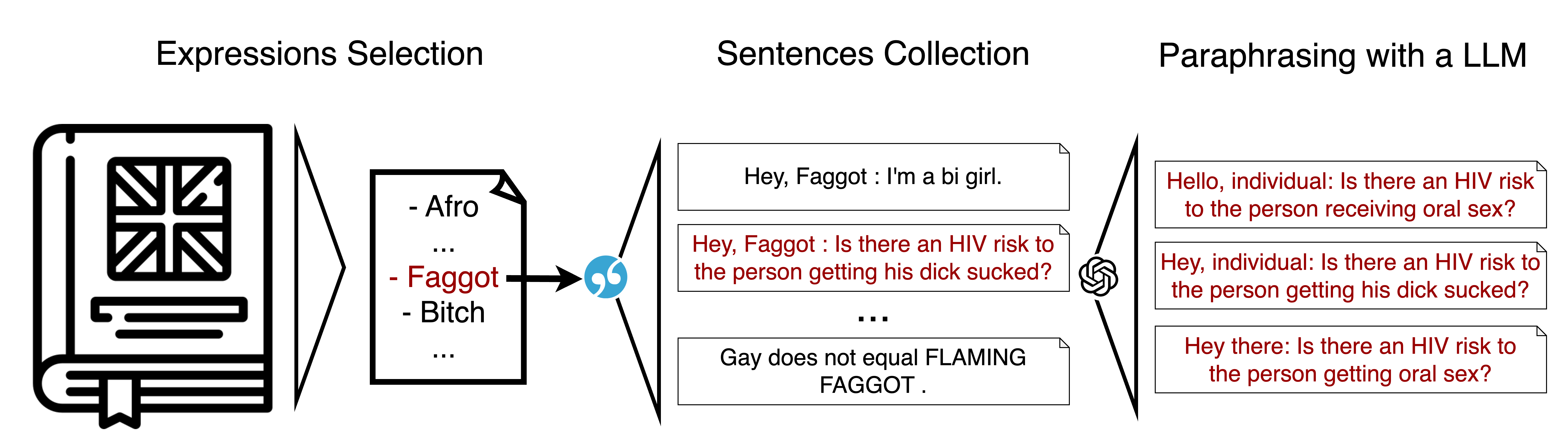}
  \caption {The Data Creation process: expressions are selected from specialized sources; sentences containing the selected expressions are collected from Sketch Engine; three paraphrases of each sentence are generated with an LLM.}
    \label{fig:data_creation}
\end{figure*}

The first step was devoted to the construction of the dataset. Our linguistic experts manually collected phrases, words, and terms (hereafter referred to as ``expressions'') spanning different sensitivity categories and levels; then, for each expression, we gathered several sentences from an online corpus and had these sentences paraphrased multiple times by an LLM. Figure~\ref{fig:data_creation} depicts the whole dataset creation process.

\paragraph{Expressions Selection.} To ensure a comprehensive assessment of the behavior of LLMs toward sensitive language, we focused on identifying expressions that would sufficiently cover different categories, such as race, gender identity, and sexual orientation. Particular emphasis has been given to: racial slurs aimed at Black individuals, homophobic terms directed at homosexual men and homosexual women, transphobic slurs targeting transgender individuals, and misogynistic language against women. The expressions were collected from several specialized dictionaries: LGBT Lexicon Library\footnote{\url{https://lexicon.library.lgbt}}, Forbidden American English~\citep{Spears1990-ky}, the Routledge Dictionary of Modern American Slang and Unconventional English~\citep{Dalzell2018-og}, and the Oxford Dictionary of Slang~\citep{Ayto1999-yz}. These expressions and their definitions were cross-checked with additional online dictionaries, including the Oxford Learner's Dictionary\footnote{\url{https://oxfordlearnersdictionaries.com}}, Cambridge Dictionary\footnote{\url{https://dictionary.cambridge.org}}, Collins Dictionary\footnote{\url{https://collinsdictionary.com}}, and the Oxford English Dictionary\footnote{\url{https://oed.com}}. These sources offered insights into the level of impoliteness and informality of each expression, helping to guide the balance of extracted expressions based on language sensitivity. A total of 599 expressions were collected for this study.

\paragraph{Sentences Collection.} We employed Sketch Engine\footnote{\url{https://sketchengine.eu}}, a popular online tool for the analysis of text corpora, to gather sentences containing the collected expressions. In particular, we employed the concordance tool through the Sketch Engine JSON API, and downloaded sentences from the English Web Corpus (enTenTen21,~\citealp{tenten-corpora}). To ensure enough context was provided within the sentence, we controlled the retrieval process by filtering for those sentences having a length between 3 and 40 words. Then, for each of the 599 expressions, we collected at most 45 sentences, resulting in a total of 23,347 sentences. Here is an example illustrating one of the collected sentences:
\begin{enumerate}
    \item[(S1)] \texttt{Hey, Faggot : Is there an HIV risk to the person getting his dick sucked?}
\end{enumerate}

\paragraph{Paraphrasing with an LLM.} For each sentence collected in the previous step, (hereafter referred to as the ``original sentence''), we used an LLM to generate paraphrases (hereafter referred to as the ``paraphrased sentences''). Specifically, we employed GPT-4o-mini-2024-07-18~\citep{gpt-4o-mini-article}, accessed via API, and prompted with the following instruction:

\begin{quote}
    \textit{Paraphrase the following sentence:} \texttt{\{original\_sentence\}.}
\end{quote}

No additional guidance on style or content was provided, and we did not instruct the model to tone down or alter any offensive language. Our goal was to observe the model’s default behavior. To introduce variability, we paraphrased each original sentence three times, yielding a total of 70,041 paraphrased sentences. Notably, GPT-4o-mini, like its larger counterparts, is designed to avoid generating disallowed content~\citep{openai2024gpt4technicalreport}. In cases where an original sentence contained extremely offensive language, the model occasionally refused to paraphrase it, responding with a warning or refusal. However, these instances were rare, representing a minority, with only 232 direct refusals out of a total of 70,041 paraphrases. As an example, here are three paraphrases for the sentence S1:
\begin{enumerate}
    \item[(P1)] \texttt{Hello, individual: Is there an HIV risk to the person receiving oral sex?}
    \item[(P2)] \texttt{Hey, individual: Is there an HIV risk to the person getting his dick sucked?}
    \item[(P3)] \texttt{Hey there: Is there an HIV risk to the person getting oral sex?}
\end{enumerate}

\subsection{Sentence Sensitivity Annotation}
\label{sec:annotation}

Given the dataset of collected original sentences and the generated paraphrased sentences, we aimed to evaluate how paraphrasing affected the sensitivity of the text. To measure the sensitivity levels, we designed a classification schema with four sensitivity categories, ranging from less sensitive (\textit{Formal/Polite} and \textit{Informal}) to more sensitive (\textit{Derogatory} and \textit{Taboo}). This schema was partially guided by insights from the online dictionaries referenced earlier. The categories are defined as follows:

\begin{itemize}
    \item \textbf{Formal/Polite}: sentences that contain very respectful or refined language (e.g., honorifics, courteous language).
    \item \textbf{Informal}: casual or colloquial sentences that are not offensive, but very conversetional or relaxed (e.g., slang, dialect grammar).
    \item \textbf{Derogatory}: sentences containing insulting or demeaning language (e.g., slurs, name-calling, provocative language) and targeting individuals or groups.
    \item \textbf{Taboo}: sentences using a profane or obscene language, making use of forbidden terminology, that is generally considered highly offensive or vulgar.
\end{itemize}

This classification schema was applied by expert annotators and automated approaches to assign a sensitivity category to both original and paraphrased sentences.

\paragraph{Annotations by Human Experts.} To obtain a reliable estimate of the effect of paraphrasing on sentence sensitivity (RQ1), we had human experts annotate a subset of the collected dataset. We recruited 10 native English speakers with research or teaching experience in the field of linguistics. Annotators were provided with detailed guidelines, which illustrated the scope of the annotation process, and informed them about the usage of sensitive language. They also had access to definitions and examples corresponding to the four sensitivity categories of our classification schema. Additionally, we allowed them to opt for a \textit{Difficult to say} option in case a given sentence lacked sufficient context for a clear classification.

We sampled 1250 original sentences, uniformly distributed with respect to expressions. For each original sentence, we randomly selected one of its three paraphrases, resulting in a total of 2500 sentences. Each annotator was assigned a batch of 500 sentences: within each batch, both the original sentence and its corresponding paraphrased version appeared. To ensure consistency, each sentence pair was also repeated in the batch of another randomly selected annotator. As a result, each sentence, whether original or paraphrased, received two independent annotations. 

To avoid any influence, the experts were unaware of whether a given sentence was original or paraphrased, and no explicit indication was provided regarding the sentence’s pair. In total, we collected annotations for 2269 sentences. These annotations serve both as indicators of how paraphrasing affected the sentence sensitivity, as well as ground truth for evaluating automated approaches.

\paragraph{Annotations by Automated Approaches} To evaluate how closely LLMs' sensitivity perceptions align with humans (RQ2), we tested the capabilities of automatic classifiers in assigning sensitivity categories to the set of 2269 annotated sentences. We evaluate the performances of these classifiers against the ground truth elaborated by the annotators in the previous step. In particular, we consider two classes of automated classifiers: open-source LLMs and traditional text classifiers. 

Prior research indicated that open-source LLMs can achieve competitive results on text classification, even when prompted in a zero-shot manner~\citep{open-source-llm-annotation}. For this reason, we employed some of the most recent and capable small-sized open-source models. We intentionally avoided the involvement of a proprietary model to ensure that our approach could be reproduced without requiring access to proprietary systems, but also to avoid potential confirmation bias led by re-using the same model employed for paraphrasing, i.e. GPT-4o-mini. For the experiments, we used Llama 3.1 8B~\citep{grattafiori2024llama3herdmodels}, Gemma 2 9B~\citep{gemmateam2024gemma2improvingopen}, Phi 3.5 Mini~\citep{abdin2024phi3technicalreporthighly}, Ministral 8B~\citep{mistralMinistralMinistraux}, OLMo 2 13B~\citep{olmo20252olmo2furious}, and Qwen 2.5 14B~\citep{qwen2025qwen25technicalreport}. All the LLMs were prompted in the following manner:


\begin{quote}
    \textit{Classify the sensitivity of the provided sentence. Choose exactly one class from the following list: Formal/Polite, Informal, Derogatory, Taboo. Provide your classification as a single word only (without any additional text).\\
    Sentence for Classification:}
    \texttt{\{sentence\}}
\end{quote}

We also tested the abilities of traditional text classifiers in predicting the sensitivity categories of sentences. The text was preprocessed using a Count Vectorizer, and the dataset was split into an 80\% training set and a 20\% test set. Each model’s hyperparameters were tuned using grid search and cross-validation. The classifiers employed were: Nearest Neighbors~\citep{nn-classifier}, Linear SVM~\citep{Cortes1995-wt}, Random Forest~\citep{Breiman2001-qf}, Naïve Bayes~\citep{Hand2001-oz}, Multi-Layer Perceptron~\citep{Rumelhart1986-to}, and Born Classifier~\citep{NEURIPS2022_c88d0c9b}. 

\begin{table*}[ht!]
\small
\centering
\begin{tabularx}{\textwidth}{@{}lcYYYc@{}}
\toprule
\textbf{Model}    & \textbf{$\Delta$ F/P} & \textbf{$\Delta$ I} & \textbf{$\Delta$ D} & \textbf{$\Delta$ T} & \textbf{$\Delta$MSD} \\ \midrule
Human Experts     & $0.16\;$\small{$(0.46)$}       & $-0.38\;$\small{$(0.59)$}    & $\mathbf{-0.87}\;$\small{$(0.90)$}    & $\mathbf{-1.7}\;$\small{$(0.94)$}    & -    \\ \midrule
\multicolumn{5}{c}{\textit{Open-Source Large Language Models}}                              \\ \midrule
Llama 3.1 8B~\citeyearpar{grattafiori2024llama3herdmodels}      & $0.05\;$\small{$(0.23)$}       & $-0.23\;$\small{$(0.48)$}    & $\mathbf{-0.52}\;$\small{$(0.65)$}    & $\mathbf{-2.33}\;$\small{$(1.15)$}    & $0.05$    \\
Gemma 2 9B~\citeyearpar{gemmateam2024gemma2improvingopen}        & $0.12\;$\small{$ (0.46)$}       & $\mathbf{-0.54}\;$\small{$ (0.75)$}    & $\mathbf{-1.18}\;$\small{$ (1.11)$}    & $\mathbf{-1.34} \;$\small{$(1.32)$}    & {\ul{$0.04$}}    \\
Phi 3.5 Mini~\citeyearpar{abdin2024phi3technicalreporthighly}      & $0.21 \;$\small{$(0.43)$}       & $0.03\;$\small{$ (0.38)$}     & $-0.32\;$\small{$ (0.71)$}    & $\mathbf{-0.90}\;$\small{$ (0.98)$}    & $0.16$     \\
Ministral 8B~\citeyearpar{mistralMinistralMinistraux}      & $0.14\;$\small{$ (0.51)$}       & $\mathbf{-0.86}\;$\small{$ (0.38)$}    & $\mathbf{-0.83}\;$\small{$ (1.00)$}    & $\mathbf{-2.00}\;$\small{$ (1.15)$}    & $0.08$     \\
OLMo 2 13B~\citeyearpar{olmo20252olmo2furious}        & $0.24\;$\small{$ (0.57)$}       & $-0.21\;$\small{$ (0.59)$}    & $-0.34\;$\small{$ (0.67)$}    & $\mathbf{-0.70}\;$\small{$ (0.86)$}    & $0.11$     \\
Qwen 2.5 14B~\citeyearpar{qwen2025qwen25technicalreport}      & $0.04\;$\small{$ (0.20)$}       & $-0.17\;$\small{$ (0.67)$}    & $-0.38\;$\small{$ (0.89)$}    & $\mathbf{-0.60}\;$\small{$ (0.96)$}    & $0.10$     \\ \midrule
\multicolumn{5}{c}{\textit{Traditional Text Classifiers}}                                   \\ \midrule
Nearest Neighbors~\citeyearpar{nn-classifier} & $\mathbf{0.52}\;$\small{$ (0.64)$}      & $-0.23\;$\small{$ (0.54)$}    & $\mathbf{-1.00}\;$\small{$ (0.63)$}     & -    & $0.06$               \\
Linear SVM~\citeyearpar{Cortes1995-wt}        & $0.35\;$\small{$ (0.64)$}       & $-0.24\;$\small{$ (0.66)$}    & $\mathbf{-1.00}\;$\small{$ (0.83)$}     & $\mathbf{-2.14}\;$\small{$ (0.69)$}    & $0.02$    \\
Random Forest~\citeyearpar{Breiman2001-qf}     & $0.05\;$\small{$ (0.21)$}       & $-0.34\;$\small{$ (0.48)$}    & $\mathbf{-0.67}\;$\small{$ (0.58)$}    & -    & $0.02$               \\
Naïve Bayes~\citeyearpar{Hand2001-oz}       & $0.23\;$\small{$ (0.49)$}       & $-0.27\;$\small{$ (0.56)$}    & $\mathbf{-0.82}\;$\small{$ (0.87)$}    & $\mathbf{-2.00}\;$\small{$ (0.00)$}    & {\ul{$0.01$}}      \\
MLP~\citeyearpar{Rumelhart1986-to}               & $0.32\;$\small{$ (0.62)$}       & $-0.34\;$\small{$ (0.52)$}   & $\mathbf{-0.87}\;$\small{$ (0.76)$}    & $\mathbf{-1.67}\;$\small{$ (1.03)$}    & {\ul{$0.01$}}    \\
Born Classifier~\citeyearpar{NEURIPS2022_c88d0c9b}   & $0.29\;$\small{$ (0.61)$}       & $-0.29\;$\small{$ (0.58)$}    & $\mathbf{-0.72}\;$\small{$ (0.89)$}    & $\mathbf{-1.42}\;$\small{$ (1.00)$}    & $0.02$    \\ \bottomrule
\end{tabularx}

\caption{Sensitivity shift ($\Delta$) is measured for each level using annotations from Human Experts, Open-Source LLMs, and Traditional Classifiers, with standard deviations in parentheses. The last column ($\Delta$MSD) shows the mean squared difference from expert annotations, with models closest to experts underlined.}
\label{tab:deltas}
\end{table*}

\subsection{Evaluation Methods}
\label{sec:eval}
In this section, we describe the evaluation methods used to compare the sensitivity levels of original sentences and their paraphrases. The insights gained from these methods will help us address our research questions.

\paragraph{Detecting Sensitivity Shift.} The examination of potential sensitivity shifts between original and paraphrased sentences is conducted with the employment of confusion matrices. In particular, using the provided sensitivity classification schema, we construct a $4 \times 4$ confusion matrix, with rows representing the categories of original sentences and columns representing the categories of paraphrased sentences. Thus, each cell $(i,j)$ in the matrix counts how many original sentences in category $i$ were labeled as category $j$ after paraphrasing. If GPT-4o-mini perfectly preserved the sensitivity level of sentences, we would expect most counts to fall on the diagonal, i.e. no change in sensitivity. Deviations from the diagonal indicate shifts or movements in sensitivity classification due to paraphrasing. To statistically test the presence of sensitivity shifts, we employed Bowker’s test~\citep{Bowker1948-ym}, whose null hypothesis tests that the contingency table is symmetric around the main diagonal.

\paragraph{Measuring Sensitivity Shift.} To assess how paraphrasing affects the perceived sensitivity of a sentence, we define the average sensitivity shift metric $\Delta$. Let $S = \{s_1, \dots, s_n\}$ be the set of $n$ original sentences. For each $s_i\in S$, let $P_i = \{p_{i1}, \dots, p_{im_i}\}$ be the set of $m_i$ paraphrases of $s_i$. Sensitivity is categorized into four ordinal levels, represented by the ordered set $C = \{1, \dots, 4\}$ where: 1 corresponds to \textit{Formal/Polite}; 2 corresponds to \textit{Informal}; 3 corresponds to \textit{Derogatory}; 4 corresponds to \textit{Taboo}. Each original sentence $s_i$ is assigned a sensitivity level $c_i \in C$, while each paraphrase $p_{ij}\in P_i$ is assigned a sensitivity level $\hat{c}_{ij} \in C$. Both levels are determined either by human annotators or through automated methods. To analyze the shift in sensitivity starting from a particular sensitivity level $c$, we define the set of the indices of original sentences associated with $c$ as $S_c = \{i\, |\,  c_i = c\}$. To quantify the average shift in sensitivity for sentences originally labeled with level $c$, we compute: $$\Delta_c = \frac{1}{\sum\limits_{i\in S_c}m_i}\sum_{i \in S_c}\sum_{j=1}^{m_i}(\hat{c}_{ij} - c_i).$$ This metric quantifies the average change in sensitivity between original sentences labeled with level $c$ and their corresponding paraphrases. A positive $\Delta_c$ indicates that paraphrases tend to increase in sensitivity, while a negative $\Delta_c$ suggests that paraphrasing generally makes sentences less sensitive. Consider that if $0.5 \leq |\Delta_c|<1.5$, we can conclude that, on average, the sensitivity level has shifted by one level, either higher or lower. Similarly: for $1.5 \leq |\Delta_c|<2.5$, the average shift is two levels; for $2.5 \leq |\Delta_c|\leq 3$, the average shift is three levels; if $|\Delta_c|<0.5$, the change is negligible, meaning the sensitivity level remains approximately the same. Since sensitivity levels are defined on a scale of 1 to 4, the maximum possible shift is three levels and the minimum is zero.
Lastly, to summarize how much each automated classification method deviates from expert assessments, we compute the $\Delta$ Mean Squared Difference ($\Delta$MSD) for a given automated classifier $A$ as $$\Delta \text{MSD}^A = \frac{1}{|C^A|}\sum\limits_{c_i \in C^A}(\Delta_{c_i} - \Delta^E_{c_i})^2$$ with $C^A$ representing the sensitivity levels available for a given classifier $A$, and $\Delta^E$ denoting the sensitivity shift values derived from expert annotations.  

\paragraph{Evaluating Automated Classifiers.} We evaluate the abilities of both the LLMs, prompted in a zero-shot manner, and the traditional methods in classifying sentences according to their sensitivity level against the expert labels. The evaluation is conducted by computing overall accuracy and $F_1$ score, as well as the $F_1$ for each sensitivity category.

\section{Results}
\label{sec:results}

In this section we provide insights on the paraphrasing effect of GPT-4o-mini, comparing the expert annotations of original sentences versus paraphrased sentences, and evaluate the performances of automatic classifiers in aligning their sensitivity levels with those of human experts.

\subsection{Sensitivity Shifts in Expert Annotations}
\label{sec:rq1_res}

\begin{figure}[ht!]
    \centering
    \includegraphics[width=0.8\linewidth]{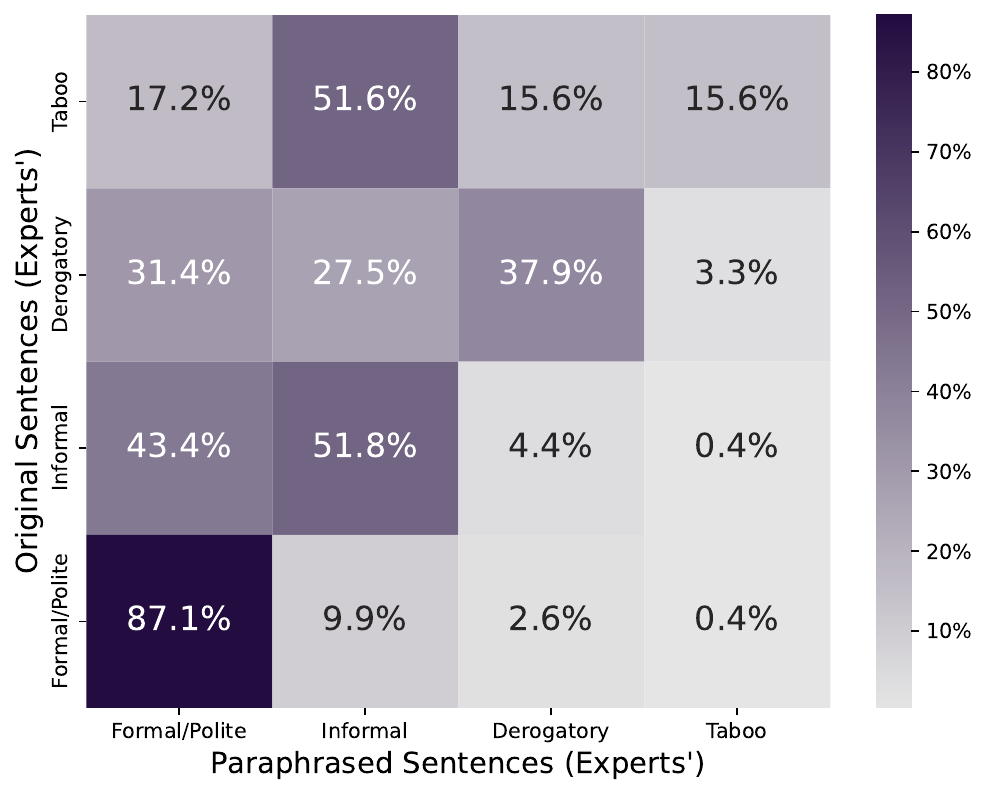}
    \caption{The confusion matrix shows sensitivity classes (Original vs. Paraphrased) based on expert annotations. Rows represent original sentence categories, columns represent paraphrased categories, and values show the percentage of sentences that shifted between categories.}
    \label{fig:confmatrix}
\end{figure}

We firstly explore whether the paraphrases generated by GPT-4o-mini systematically shift the sensitivity classifications compared to the original sentences. Figure~\ref{fig:confmatrix} shows the confusion matrix of expert annotations for the selected subset of annotated sentence pairs.

\begin{figure*}[ht!]
    \centering
    \includegraphics[width=0.9\linewidth]{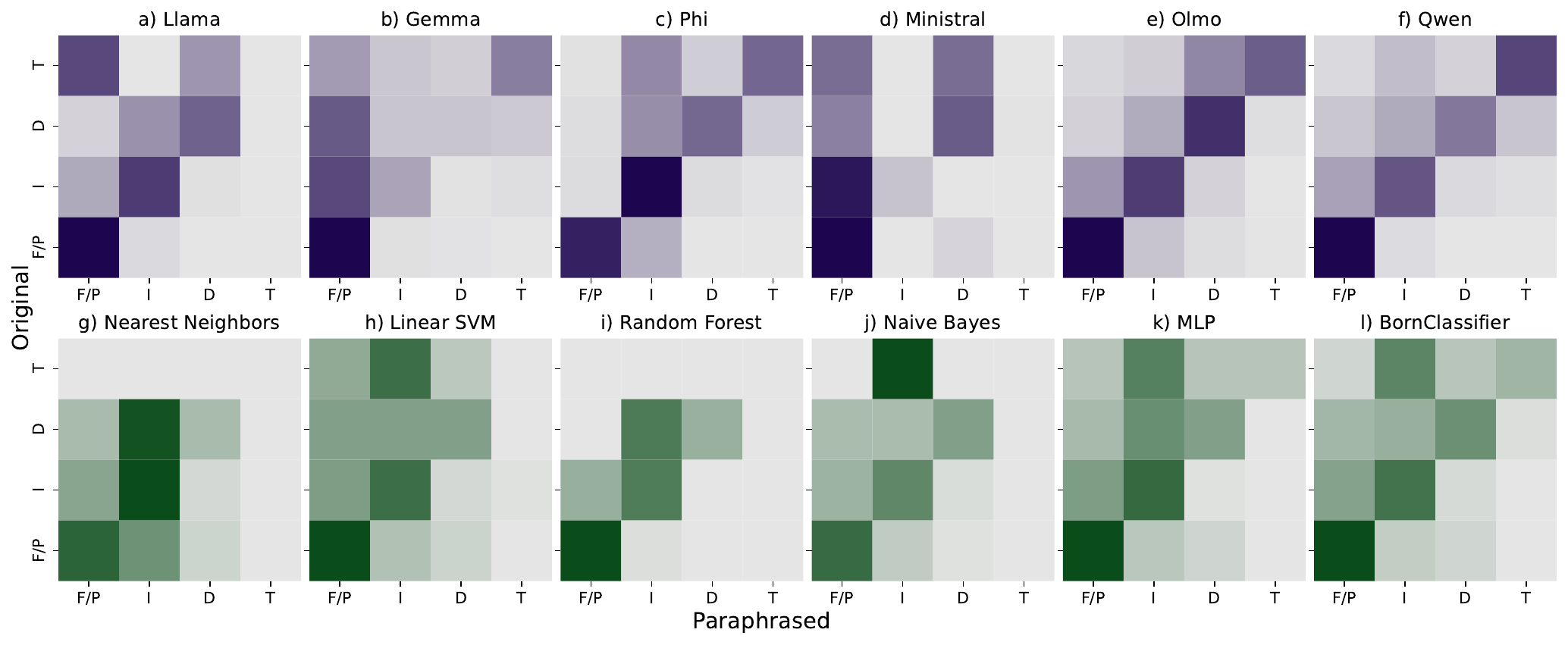}
    \caption{Confusion matrices show sensitivity classifications (Original vs. Paraphrased) for Open-Source LLMs (purple) and traditional classifiers (green). Rows indicate original sentence classifications, columns show paraphrased classifications, and color intensity reflects the percentage of category shifts, with stronger colors indicating higher transition rates.}
    \label{fig:automated-confmatrix}
\end{figure*}

The diagonal cells, which represent identical sensitivity classifications before and after paraphrasing, show that paraphrasing preserves sensitivity best for less sensitive categories (\textit{Formal/Polite} $87.1\%$, \textit{Informal} $51.8\%$) with respect to highly sensitive ones (\textit{Derogatory} $37.9\%$, \textit{Taboo} $15.6\%$), suggesting that sensitivity is rarely added when it was originally absent. The higher values in the top-left corner, compared to the lower values in the lower-right corner, are evidence of a general tendency of paraphrasing toward reducing sensitivity: the majority ($51.6\%$) of \textit{Taboo} original sentences were paraphrased as \textit{Informal}, while most ($58.9\%$) of originally \textit{Derogatory} were paraphrased as \textit{Formal/Polite} or as \textit{Informal}. These results are confirmed by Bowker’s test statistic, ($\chi^2=138.255$, $df=6$, $p<0.0001$), which strongly rejects the null hypothesis of symmetry.

To evaluate the magnitude of the sensitivity shift observed in expert annotations, we compute $\Delta$ for each sensitivity class. The first row (\textit{Human Experts}) in Table~\ref{tab:deltas} illustrates the resulting $\Delta$ based on human annotations. Considering the negative sign of $\Delta$ for higher levels of sensitivity, we can confirm that there is indeed a sensitivity shift toward less sensitive levels when GPT-4o-mini is asked to paraphrase. The shift is particularly large for original sentences classified as \textit{Taboo}, with a $\Delta$ of $-1.7$, which is still significant even considering values within one standard deviation.


\subsection{Automated Classification Performance}
\label{sec:rq2_res}

We address the second research question evaluating how well local automated classifiers align with experts in annotating sentence sensitivity. 

Figure~\ref{fig:automated-confmatrix} depicts confusion matrices resulting from automated classifier annotations. Comparing these matrices to Figure~\ref{fig:confmatrix}, created with experts' annotations, we can perceive a closer overlap of some matrices resulting from traditional text classifiers, in particular those from MLP and BornClassifier. On the contrary, many LLMs' confusion matrices have higher values on the main diagonal (e.g., Phi, OLMo, Qwen). This suggests that these LLMs perceive GPT-4o-mini paraphrases with the same level of sensitivity as the original sentences, refusing the hypothesis of paraphrasing affecting sensitivity level. Other LLMs (e.g. Gemma and Ministral) appear to have a tendency to classify paraphrases as \textit{Formal/Polite}, regardless of the original sentences' sensitivity level. Lastly, LLaMA tends to classify as \textit{Formal/Polite} paraphrases of original sentences labeled as \textit{Taboo}, and in general, it does not classify paraphrases within the \textit{Taboo} sensitivity category.

\begin{table*}[ht!]
\small
\centering
\begin{tabularx}{\textwidth}{@{}lYYYYYY@{}}
\toprule
\textbf{Model}    & \textbf{Accuracy}   & \multicolumn{5}{c}{\textbf{F$_1$ Score}}                                            \\
                  & Overall             & Formal/Polite & Informal      & Derogatory    & Taboo         & Overall       \\ \midrule
\multicolumn{7}{c}{\textit{Open-Source Large Language Models}}                                                          \\ \midrule
LLama 3.1 8B~\citeyearpar{grattafiori2024llama3herdmodels}      & $0.43$                & $0.45$          & $0.45$          & $0.39$          & $0.05$          & $0.42$          \\
Gemma 2 9B~\citeyearpar{gemmateam2024gemma2improvingopen}        & {\ul $0.52$}          & {\ul $\mathbf{0.68}$} & $0.35$          & $0.17$          & {\ul $0.35$}    & {\ul $0.47$}    \\
Phi 3.5 Mini~\citeyearpar{abdin2024phi3technicalreporthighly}      & $0.35$                & $0.10$          & $0.49$          & $0.36$          & $0.26$          & $0.28$          \\
Ministral 8B~\citeyearpar{mistralMinistralMinistraux}      & $0.45$                & $0.63$          & $0.03$          & {\ul $0.40$}    & $0.05$          & $0.37$          \\
OLMo 2 13B~\citeyearpar{olmo20252olmo2furious}        & $0.42$                & $0.43$          & $0.45$          & $0.39$          & $0.23$          & $0.42$          \\
Qwen 2.5 14B~\citeyearpar{qwen2025qwen25technicalreport}      & $0.42$                & $0.49$          & {\ul $0.59$}    & $0.31$          & $0.25$          & $0.45$          \\ \midrule
\multicolumn{7}{c}{\textit{Traditional Text Classifiers}}                                                               \\ \midrule
Nearest Neighbors~\citeyearpar{nn-classifier} & $0.40$                & $0.13$          & $0.49$          & $0.43$          & $0.00$          & $0.37$          \\
Linear SVM~\citeyearpar{Cortes1995-wt}        & $0.52$                & $0.20$          & $0.67$          & $0.45$          & $0.33$          & $0.49$          \\
Random Forest~\citeyearpar{Breiman2001-qf}     & $0.46$                & $0.06$          & $0.62$          & $0.33$          & $0.00$          & $0.38$          \\
Naïve Bayes~\citeyearpar{Hand2001-oz}       & $0.49$                & $0.23$          & $0.63$          & $0.46$          & $0.10$          & $0.47$          \\
MLP~\citeyearpar{Rumelhart1986-to}               & {\ul $\mathbf{0.55}$} & $0.30$          & {\ul $\mathbf{0.69}$} & {\ul $\mathbf{0.48}$} & $0.31$          & $0.52$          \\
Born Classifier~\citeyearpar{NEURIPS2022_c88d0c9b}   & $0.54$                & {\ul $0.32$}    & $0.66$          & $0.46$          & {\ul $\mathbf{0.57}$} & {\ul $\mathbf{0.53}$} \\ \bottomrule
\end{tabularx}
\caption{Classification performance of Open-Source LLMs and Traditional Text Classifiers in predicting sentence sensitivity. Best in bold; best in model category underlined.}
\label{tab:classification-performance}
\end{table*}

Regarding the magnitude of the sensitivity shift perceived by automated classifiers, the rows under \textit{Open-Source LLMs} and \textit{Traditional Classifiers} in Table~\ref{tab:deltas} report the $\Delta$ for each sensitivity class, based on annotations from each automated classifier. Among LLMs, Gemma is the most similar to Human Experts in terms of $\Delta$ (lowest $\Delta$MSD), meaning that it has a similar perception of sensitivity as experts' in classifying original and paraphrased sentences. Overall, LLMs consistently agree on the sign of $\Delta$, reinforcing GPT-4o-mini’s tendency to sanitize paraphrases in terms of sensitive content, with the exception of Phi. Conforming with our earlier observations about confusion matrices, some LLMs tend to assign the same sensitivity class to both original and paraphrased sentences, as reflected in the lower $\Delta$ values.
Among the six \textit{Traditional Text Classifiers}, their $\Delta$MSD values indicate a closer alignment with expert annotations compared to \textit{Open-Source LLMs}. Notably, the Naïve Bayes and MLP classifiers achieve the lowest $\Delta$MSD values among all methods. These classifiers exhibit a stronger tendency toward assigning less sensitive levels to GPT-4o-mini paraphrases, as evidenced by higher $\Delta$ scores for both \textit{Taboo} and \textit{Derogatory} sensitivity levels, empirically suggesting greater adherence to expert classifications.

Table~\ref{tab:classification-performance} summarizes the performances of both the six representative open-source LLMs used in a zero-shot classification setting and the six traditional text classifiers. The performances are expressed in terms of accuracy and $F_1$ score computed for each sensitivity level and overall. Traditional methods outperform LLMs overall, particularly in informal and derogatory language. The classical MLP and Born Classifier consistently ranked highest across multiple categories, with the latter achieving the highest overall $F_1$ score of $0.53$. However, the overall classification performances across both open-source LLMs and traditional text classifiers remain relatively poor, exhibiting a failure of automated systems in aligning with the human perception of sensitive language.

\section{Limitations}
\label{sec:limitation}
In this section, we discuss the main limitations of this work. First, regarding dataset creation, we relied solely on the English Web Corpus as our source for collecting sentences, which could have limited the scope of our evaluation given the specificity of this data source. Additionally, despite being experts in linguistics and native English speakers, annotators’ inherent preconceptions may have influenced their judgments. Moreover, a higher number of annotations would benefit the robustness of the discussed results. To enhance annotation quality and the overall insights' relevance, we plan to expand the annotation task to a larger and more diverse group of annotators in the future and to include a larger variety of data sources.

\section{Conclusion}
\label{sec:conclusion}
This work empirically explores the implicit sanitization performed by GPT-4o-mini when paraphrasing sensitive content. Our findings show that GPT-4o-mini systematically reduces the sensitivity level of paraphrased text, suggesting that alignment techniques cause an implicit moderation behavior even without explicit detoxification instructions. These results confirm prior observations of style transfer in LLMs, where offensive or explicit content is reduced by design to minimize harm~\citep{openai2024gpt4technicalreport}. Also, local LLMs (in zero-shot mode) slightly underperform compared to traditional text classifiers when matching experts' classes. Direct supervision with in-domain data boosts performance, highlighting that generic LLMs may not be a quick substitute for trained classifiers in risky moderation tasks. Simultaneously, open-source LLMs remain attractive for scalability, rapid deployment, and easier prompt customization.
These insights deepen our understanding of how aligned LLMs sanitize user input and underscore the importance of careful oversight when automating moderation at scale.


\bibliography{custom}




\end{document}